\pdfoutput=1
\documentclass[11pt]{article}

\usepackage{acl}

\usepackage{times}
\usepackage{latexsym}
\usepackage{tikz}
\usepackage{multirow}
\usepackage{multicol}
\usepackage{booktabs}
\usepackage{graphicx}
\usepackage[T1]{fontenc}
\usepackage[utf8]{inputenc}

\usepackage{microtype}

\usepackage{inconsolata}
\newcommand\blfootnote[1]{%
  \begingroup
  \renewcommand\thefootnote{}\footnote{#1}%
  \addtocounter{footnote}{-1}%
  \endgroup
}
\title{MasonTigers at SemEval-2024 Task 8: Performance Analysis of Transformer-based Models on Machine-Generated Text Detection}

\author{Sadiya Sayara Chowdhury Puspo\textsuperscript{*},  Md Nishat Raihan\textsuperscript{*}, Dhiman Goswami\textsuperscript{*},\\ {\bf  Al Nahian Bin Emran, Amrita Ganguly, \"Ozlem Uzuner} \\ George Mason University, USA \\
 \texttt{spuspo@gmu.edu}
 }

\begin{document}
\maketitle
\begin{abstract}
This paper presents the \textit{MasonTigers'} entry to the SemEval-2024 Task 8 - Multigenerator, Multidomain, and Multilingual Black-Box Machine-Generated Text Detection. The task encompasses Binary Human-Written vs. Machine-Generated Text Classification (Track A), Multi-Way Machine-Generated Text Classification (Track B), and Human-Machine Mixed Text Detection (Track C). Our best performing approaches utilize mainly the ensemble of discriminator transformer models along with sentence transformer and statistical machine learning approaches in specific cases. Moreover, zero-shot prompting and fine-tuning of FLAN-T5 are used for Track A and B. \blfootnote{\bf * denotes equal contribution.}
\end{abstract}

\section{Introduction}

In academia and beyond, machine-generated content is proliferating across news platforms, social media, forums, educational materials, and scholarly works. Breakthroughs in large language models (LLMs), like GPT-3.5 and GPT-4, facilitate the creation of fluent responses to diverse user queries. While this capability raises prospects of replacing human labor in various tasks, concerns arise about potential misuse, including the generation of deceptive misinformation \cite{chen2023combating} and completing student assignments, which hinders the development of essential writing skills \cite{jungherr2023using}.%It is crucial to underscore the need for careful observation and proactive measures. It is noteworthy that humans only slightly outperform random chance in distinguishing between machine-generated and human-written text. 
This highlights the importance of developing automated systems to detect and mitigate the potential abuse of machine-generated content, as well as distinguishing between machine-written and human-generated text. Additionally, 
Prior studies (ZeroGPT \footnote{\url{www.zerogpt.com/}}; \citeauthor{mitchell2023detectgpt}, \citeyear{mitchell2023detectgpt}; \citeauthor{bao2023fast}, \citeyear{bao2023fast}) predominantly adopted a binary classification approach for machine-generated text (MGT), with a primary focus on English. However, there has been limited research addressing the amalgamation of human-written and MGT texts \cite{wang2024m4}.

In response to these limitations, SemEval-2024 introduces a shared task: Multigenerator, Multidomain, and Multilingual Black-Box Machine-Generated Text Detection \cite{semeval2024task8}. This task comprises three subtasks, each targeting different aspects of machine-generated text complexity.Subtask A focuses on Binary Human-Written vs. MGT Classification, involving two tracks: monolingual and multilingual. Subtask B tackles Multi-Way Machine-Generated Text Classification to identify the source of a given text. Subtask C involves detecting the transition point within a mixed text, determining where it shifts from human-written to machine-generated. The data provided for this task is an expansion of the M4 dataset \cite{wang2024m4} and benchmark evaluation of \cite{wang2024m4gt}.
%Subtask A focuses on Binary Human-Written vs. MGT Classification, where the goal is to determine whether a given full text is authored by a human or generated by a machine, comprising of two tracks: a monolingual and multilingual. Subtask B delves into Multi-Way Machine-Generated Text Classification, aiming to identify the source of a given full text—whether it originated from a human writer or a specific language model. Subtask C involves detecting the transition point within a mixed text, where the initial segment is human-written and the subsequent part is machine-generated. The task is to determine where the text changes from human-written to machine-generated.
.

In conducting these tasks, we conduct a range of experiments and observe that ensemble methods outperform individual models significantly in classification tasks, e.g., \citet{goswami2023nlpbdpatriots} \citet{emran2024masonperplexity}, \citet{ganguly2024masonperplexity}. Our weighted ensemble approaches achieve accuracies of 74\%, 60\% and 65\% in subtask A monolingual; multilingual tracks and subtask B respectively, given that we have used different models for both tasks.  In subtask C, we explore different setups, ensembling which results in Mean Absolute Error (MAE) of 60.78. For the classifications, we utilize zero-shot prompting and fine-tuning of FlanT5\footnote{\url{huggingface.co/google/flan-t5-base/}}, while adhering to the restriction of no data augmentation in this task.
%In the monolingual track of subtask A, combining Roberta \cite{liu2019roberta}, DistilBERT \cite{sanh2019distilbert}, and ELECTRA \cite{clark2020electra}, a weighted ensemble achieves an accuracy of 74\%.  Similarly, in the multilingual track, an ensemble of LASER \cite{li2020transformer}, mBERT \cite{devlin2018bert}, and XLMR \cite{goyal2021larger} achieves accuracy of 60\%.  
%In subtask B, combining, weighted ensemble of Roberta, ELECTRA, Deberta \cite{he2020deberta} in an ensemble achieves an accuracy of 65\%. In subtask C, we explore different setups using Linear Regression \cite{gross2003linear} and ElasticNet \cite{zou2005regularization} resulting in a Mean Absolute Error (MAE) of 60.78. 

\begin{table*}[!h]
\centering
\begin{tabular}{l|cccc|c|c|c}
\hline
\multicolumn{1}{c|}{\textbf{Source}} &
\multicolumn{5}{c|}{\textbf{Train}} & \multicolumn{2}{c}{\textbf{Dev}}\\
\cline{1-8}
  & \textbf{chatGPT} & \textbf{Cohere} & \textbf{Davinci} & \textbf{Dolly} & \textbf{Human}  & \textbf{Bloomz} & \textbf{Human}\\
\hline
arxiv & 3000 & 3000 & 2999 & 3000 & 15498 & 500 & 500 \\
peerread & 2344 & 2342 & 2342 & 2344 & 2357 & 500 & 500\\
reddit & 3000 & 3000 & 3000 & 3000 & 15500 & 500 & 500\\
wikihow & 3000 & 3000 & 3000 & 3000 & 15499 & 500 & 500\\
wikipedia & 2995 & 2336 & 3000 & 2702 & 14497 & 500 & 500\\
\hline
\bf Total  & & & & \bf 54406 & \bf 63351 & \bf 2500 & \bf 2500 \\
\hline
\end{tabular}
\caption{Label Distribution of Train and Validation Data for Binary Human-Written vs. Machine-Generated Text Classification (Subtask A - Monolingual)}
\label{tab:mono1}
\end{table*}

\section{Related Works}

The difficulties of separating human-written text from large language models and the significance of trustworthy methods for evaluation are highlighted by recent research (e.g. \citeauthor{chaka2024reviewing} \citeyear{chaka2024reviewing}, \citeauthor{elkhatat2023evaluating} \citeyear{elkhatat2023evaluating}). In terms of human evaluation of MGT, \citet{guo2023close} indicates that generated texts from large language models tend to exhibit less emotional and objective content compared to human-written texts. \citet{tang2023science} suggests that distinct signals left in the machine-generated text may facilitate the identification of suitable features to differentiate between MGT and human-written texts. Whereas, \citet{sadasivan2023aigenerated} observes that detection techniques become less effective as language models improve. Moreover, \citet{ippolito2019automatic} advocates for the importance of using both human and automatic detectors to assess the humanness of text generation systems.

Previous work in determining MGT from human-written ones include higher order n-grams \cite{gallé2021unsupervised}, utilizing linguistic patterns \cite{munoz2023contrasting}, curvature-based criterion \cite{mitchell2023detectgpt}, tweaking with multiple variables \cite{dugan2023real}, fine-tuning transformer-based models e.g., \citet{capobiancosupervised}; \citet{chen2023stadee}. Very recently,  \citet{wang2024llm} puts forward LLM-Detector, offering a fresh method for identifying text at both document and sentence levels by employing Instruction Tuning of LLMs. To tackle challenges of this field, several datasets have been released, e.g., MULTITuDE \cite{macko2023multitude}, M4 \cite{wang2024m4}.
Additionally, there have been multiple shared tasks organized related to this topic  (\citeauthor{Shamardina_2022}, \citeyear{Shamardina_2022}; \citeauthor{molla2023overview}, \citeauthor{molla2023overview}; \citeauthor{kashnitsky2022overview}, \citeyear{kashnitsky2022overview}. Despite several collective findings and techniques, as argued by \citet{sadasivan2023aigenerated}, there remains a critical need for the creation of reliable detection methods capable of accurately distinguishing between human and machine-generated text, a requirement essential across both English and other languages.  
%\citet{gallé2021unsupervised} utilize repeated higher-order n-grams for detecting machine-generated text. Several papers have been written to help advance the area by exploring techniques for identifying mixed writing that is written by humans and machines. %\citet{Dugan_Ippolito_Kirubarajan_Shi_Callison-Burch_2023} higlights the impact of multiple variables on detection performance and notes significant variance in annotator competence. \citet{krishna-etal-2022-rankgen} introduces RankGen, a model focusing on human detection of text transitions from human-written to machine-generated. Furthermore,  MULTITuDE \cite{macko2023multitude}, a benchmark dataset for multilingual machine-generated text has been introduced. This dataset addresses the scarcity of research on languages beyond English. In this area of study, there has been a recent utilization of LLMs. \citet{wang2024llm} puts forward LLM-Detector, offering a fresh method for identifying text at both document and sentence levels by employing Instruction Tuning of LLMs. The collective findings of these studies highlight diverse techniques and benchmark datasets pertinent to this common objective. 

\begin{table*}[!ht]
\centering
\scalebox{0.915}{
\begin{tabular}{l|ccccc|c|cc|c}
\hline
\multicolumn{1}{c|}{\textbf{Source}} &
\multicolumn{6}{c|}{\textbf{Train}} & \multicolumn{3}{c}{\textbf{Dev}}\\
\cline{1-10}
 & \textbf{Bloomz} & \textbf{chatGPT} & \textbf{Cohere} & \textbf{Davinci} & \textbf{Dolly} & \textbf{Human} & \textbf{ChatGPT} & \textbf{Davinci} & \textbf{Human}\\
\hline
arxiv & 3000 & 3000 & 3000 & 2999 & 3000 & 15998 &-&-&- \\
peerread & 2334 & 2344 & 2342 & 2344 & 2344 & 2857 &-&-&-\\
reddit & 2999 & 3000 & 3000 & 3000 & 3000 & 16000 &-&-&-\\
wikihow & 3000 & 3000 & 3000 & 3000 & 3000 & 15999 &-&-&-\\
wikipedia & 2999 & 2995 & 2336 & 3000 & 2702 & 14997 &-&-&-\\
Bulgarian & 0 & 3000 & 0 & 3000 & 0 & 6000 &-&-&-\\
Chinese & 0 & 2970 & 0 & 2964 & 0 & 6000 &-&-&-\\
Indonesian & 0 & 3000 & 0 & 0 & 0 & 2995 &-&-&-\\
Urdu & 0 & 2899 & 0 & 0 & 0 & 3000 &-&-&-\\
\hline
Arabic &-&-&-&-&-&-& 500 & 0 & 500\\
German &-&-&-&-&-&-& 500 & 0 & 500\\
Russian &-&-&-&-&-&-& 500 & 500 & 1000\\
\hline
\bf Total  & & & & & \bf 83571 & \bf 83846 &  & \bf 2000 & \bf 2000\\
\hline
\end{tabular}
}
\caption{Label Distribution of Train and Validation Data for Binary Human-Written vs. Machine-Generated Text Classification (Subtask A - Multilingual)}
\label{tab:mul1}
\end{table*}

\begin{table*}[!ht]
\centering
\scalebox{0.735}{
\begin{tabular}{l|cccccc|cccccc}
\hline
\multicolumn{1}{c|}{\textbf{Source}} &
\multicolumn{6}{c|}{\textbf{Train}} &
\multicolumn{6}{c}{\textbf{Dev}} \\
\cline{1-13}
 & \textbf{Bloomz} & \textbf{chatGPT} & \textbf{Cohere} & \textbf{Davinci} & \textbf{Dolly} & \textbf{Human} & \textbf{Bloomz} & \textbf{chatGPT} & \textbf{Cohere} & \textbf{Davinci} & \textbf{Dolly} & \textbf{Human} \\
\hline
arxiv & 3000 & 3000 & 3000 & 2999 & 3000 & 2998  &-&-&-&-&-&-\\
reddit & 2999 & 3000 & 3000 & 3000 & 3000 & 3000 &-&-&-&-&-&-\\
wikihow & 3000 & 3000 & 3000 & 3000 & 3000 & 2999 &-&-&-&-&-&-\\
wikipedia & 2999 & 2995 & 2336 & 3000 & 2702 & 3000 &-&-&-&-&-&-\\
\hline
peerread &-&-&-&-&-&-& 500 & 500 & 500 & 500 & 500 & 500 \\
\hline
\bf Total  & \bf 11998 & \bf 11995 & \bf 11336 & \bf 11999 & \bf 11702 & \bf 11997 & \bf 500 & \bf 500 & \bf 500 & \bf 500 &\bf 500 & \bf 500\\
\hline
\end{tabular}
}
\caption{Label Distribution of Train and Validation Data for Multi-Way Machine-Generated Text Classification (Subtask B)}
\label{tab:ds-B}
\end{table*}

\section{Datasets}
\citet{wang2024m4} collects datasets from a variety of sources, including Wikipedia (the March 2022 version), WikiHow \cite{koupaee2018wikihow}, Reddit (ELI5), arXiv, PeerRead \cite{kang2018dataset}(for English), and Baike (for Chinese). They employ web question answering for Chinese, news content for Urdu, Indonesian, and RuATD \cite{shamardina2022findings} for Russian language. The method of prompting machine-generated text (MGT) has been extensively outlined in \citet{wang2024m4}.

\begin{table}[]
\centering
\scalebox{1.1}{
\begin{tabular}{lcc}
\hline
\textbf{Label} & \textbf{ Test Data}  \\
\hline
Human (0)  & 3000  \\
chatGPT (1)  & 3000 \\
cohere (2)  & 3000 \\
davinci (3)  & 3000 \\
Bloomz (4)  & 3000 \\
Dolly (5)  & 3000\\
\hline
\bf Total & 18000\\
\hline
\end{tabular}
}
\caption{Label Distribution of Test Data for Multi-Way Machine-Generated Text Classification (Subtask B)}
\label{tab:ds-B2}
\end{table}

%To prompt MGT, the models are prompted with creating Wikipedia articles based on given titles, composing abstracts from title prompts for arXiv, generating peer reviews using title and abstract cues for PeerRead, answering questions for Reddit and Baike/Web QA, and crafting news briefs based on provided titles. Specific guidelines, such as word or token counts, are specified for particular Language Model tasks during the process by \citet{wang2024m4}.

Subtask A, Binary Human-Written vs. Machine-Generated Text Classification, in the monolingual track involves a same-domain cross-generator experiment, where instances are exclusively in English and gathered from five distinct sources with two labels: 0 and 1. Human-generated texts receive a label of 0, while machine-generated texts from four different LLMs (chatGPT, Cohere, \textit{davinci-003}, and Dolly-v2) are labeled as 1. The distribution of Train and Validation datasets, both in terms of labels and sources, along with the number of test instances, is detailed in Tables \ref{tab:mono1}. During the test phase, there are 16,272 instances labeled as 0 and 18,000 instances labeled as 1.

On the other hand, Subtask A in the multilingual track entails a cross-domain same-generator experiment. Instances are sourced from nine different sources during the training phase, including four different languages, while the validation dataset comprises three different languages as indicated in Table \ref{tab:mul1}. Similar to the monolingual task, human-generated texts are labeled as 0, and machine-generated texts from five different LLMs (Bloomz \cite{muennighoff2022crosslingual}, chatGPT, Cohere, \textit{davinci-003}, and Dolly-v2) are labeled as 1. In the test phase, there are 20,238 instances labeled as 0 and 22,140 instances labeled as 1.

Subtask B, Multi-Way Machine-Generated Text Classification, represents another cross-domain same-generator experiment. In contrast to Subtask A, Subtask C involves six labels: 0 for human, 1 for chatGPT, 2 for Cohere, 3 for \textit{davinci-003}, 4 for Bloomz, and 5 for Dolly. These labels correspond to instances sourced from five different sources. However, it's noteworthy that the sources for the training and validation data differ, and this distinction is outlined in Tables \ref{tab:ds-B} and \ref{tab:ds-B2}. %During the test phase, each label comprises 3,000 instances, resulting in a total of 18,000 instances.

Subtask C, involving Human-Machine Mixed Text Detection, provides a composite text with a human-written first part followed by a machine-generated second part. The task is to discern the boundary, and labels are provided as word indices to distinguish it. The label distribution of data is shown in Table \ref{tab:DS C}.  %There are  23\%, 3\% and 73\% instances for each training, development and test phases respectively..

\begin{table}[!ht]
\centering
\scalebox{1.1}{
\begin{tabular}{lc}
\hline
\textbf{Data} & \textbf{Count}\\
\hline
 Train & 3649\\
 Dev & 505\\
 Test & 11123\\
\hline
\end{tabular}
}
\caption{Number of Instances for Human-Machine Mixed Text Detection (Subtask C)}
\label{tab:DS C}
\end{table}

\section{Experimental Setup}

\subsection{Data Preprocessing}

In the monolingual track of subtask A, we received approximately 160K instances for training and development. To preserve the text's integrity, we eliminate special characters, extra new lines, unnecessary whitespace, and hyperlinks from the data, ensuring that only the essential text remains in subtassk A (monolingual), B \& C. However, in the multilingual track of subtask A, since none of our team members are familiar with the languages present in the instances, we only remove hyperlinks. We ensure that punctuation marks such as full stops, commas, and exclamation signs are retained in all instances, as they play a crucial role in this task \cite{tang2023science}.

\subsection{Hyperparameters}
%The monolingual track of subtask A is a binary classification task, in which, the training data for models differs from the validation phase, involving cross-generator scenarios within the same domain. While the sources of the training data remain consistent, there is a variation in the Large Language Models (LLMs) used during training and validation. 

In our experimental setup, we configure several key parameters to train our model effectively. We utilize a batch size of 16, controlling the number of training samples processed in each iteration, learning being set to 1e-5, and incorporating dropout with a rate of 0.25 to prevent overfitting by randomly dropping a fraction of units during training. Maintaining a fixed sequence length of 512 tokens ensured consistency in input data processing. For optimization, we employ the AdamW optimizer \cite{loshchilov2017decoupled}, known for its efficacy in training deep neural networks with added weight decay regularization. These experiments are conducted on a 80GB NVIDIA A100 GPU machine over the period of 24 hours, leveraging its computational power and memory capacity. By systematically adjusting these parameters, we aim to understand their influence on the model's performance, ultimately optimizing our approach for the task at hand. The adjustment of these parameters is carried out in both subtask A \& B.

\subsection{Models: SubTask A}
In monolingual track, we employ Roberta \cite{liu2019roberta}, DistilBERT \cite{sanh2019distilbert}, and ELECTRA \cite{clark2020electra}. Subsequently, we apply a weighted ensemble method, incorporating RoBERTa, DistilBERT, and ELECTRA, employing a voting strategy due to their closely comparable individual accuracies. The weights are their corresponding accuracy.

%For Notably, \citet{wang2024m4} points out the substantial dissimilarity of Bloomz data from other LLMs. Consequently, predicting text from Bloomz during the validation phase poses a challenging task for models familiar with texts generated by other LLMs. 

Similarly,  in the multilingual track, we utilize LASER \cite{li2020transformer}, mBERT \cite{devlin2018bert}, and XLMR \cite{goyal2021larger}. Following that, we deploy a weighted ensembled strategy involving these models, utilizing the voting method. 
%Additionally, we experiment with zero-shot prompting and fine-tuning using FLAN T5. Nevertheless, the sources and models for the test phase remain undisclosed for both tracks.Additionally, we explore zero-shot prompting and fine-tuning with FLAN T5 \cite{chung2022scaling}. 

\subsection{Models: SubTask B}

Subtask B, poses a considerable challenge, as opposed to the first two tracks where the model distinguishes between human and machine-generated text. Here, the model must differentiate among human-generated text and five distinct LLMs. For this, we leverage Roberta, ELECTRA, Deberta \cite{he2020deberta}, and subsequently create a weighted (weights are set as acauracy) ensemble approach of these models using voting technique. 

\subsection{Models: SubTask C}

In subtask C, we find the embedding of the training data using Term Frequency - Inverse Document Frequency (TF-IDF) \cite{aizawa2003information}, Positive Point-wise Mutual Information (PPMI) \cite{church1990word}, and the embedding using language model RoBERTa \cite{liu2019roberta}. Then for each training embedding generated by these approaches, we apply Linear Regression \cite{gross2003linear} and ElasticNet \cite{zou2005regularization} separately on these embeddings and predict the first word or index of from where the machine-generated text started in a specific data instance. We selected the word that is the starting word of the closest neighboring paragraph as the predicted word index. Then we clip the predicted values to ensure the predictions range from 0 to the length of the specific data instance (rounded if necessary). In the development phase, we find the Mean Absolute Error \cite{chai2014root} of these six predictions (three each by Linear Regression and ElasticNet). Then we perform a weighted ensemble depending on the Mean Absolute Error of the six predicted results and get our ensembled MAE in the development phase. We also perform this approach on the test data and find our smallest MAE in the evaluation phase.

\subsection{Prompting and Fine-Tuning LLM}

\begin{figure}[h]
\centering
\scalebox{.92}{
\begin{tikzpicture}[node distance=1cm]
    % Styles for nodes
    \tikzstyle{block} = [rectangle, draw, fill=green!20, text width=\linewidth, text centered, rounded corners, minimum height=4em]
    \tikzstyle{operation} = [text centered, minimum height=1em]
    % Nodes
    \node [block] (rect1) {\textbf{Prompt:}{ \textit{<Text>} If this piece of text is Human Generated, answer 0 or If Machine Generated, answer 1. }};
\end{tikzpicture}
}
\caption{Sample FlanT5 prompt.}
\label{fig:prompt1}
\end{figure}

\begin{table*}[!ht]
\centering
\scalebox{1}{
\begin{tabular}{lcc|lcc}
\hline
\multicolumn{3}{c|}{\textbf{Monolingual}} &
\multicolumn{3}{c}{\textbf{Multilingual}} \\
\hline
\textbf{Model}  & \textbf{Dev} & \textbf{Test} & \textbf{Model}  & \textbf{Dev} & \textbf{Test}\\
\hline
\bf Baseline (RoBERTa) & 0.74 & 0.88 & \bf Baseline (XLM-R) & 0.72 & 0.81\\
\hline
FLAN-T5 Prompting & 0.49 & 0.52 & FLAN-T5 Prompting & 0.42 & 0.39\\
FLAN-T5 Fine-tuning & 0.57 & 0.53 & FLAN-T5 Fine-tuning & 0.48 & 0.43\\
\hline
RoBERTa  & 0.70 & 0.73 & LASER  & 0.52 & 0.50\\
DistilBERT  & 0.69 & 0.70 & mBERT  & 0.57 & 0.58\\
ELECTRA & 0.78 & 0.71 & XLMR & 0.61 & 0.59\\
\hline
\bf Ensemble (Wt. accuracy) & 0.79 & 0.74 & \bf Ensemble (Wt. accuracy) & 0.63 & 0.60\\
\hline
\end{tabular}
}
\caption{Accuracy of Binary Human-Written vs. Machine-Generated Text Classification (Subtask A)}
\label{tab:trackA_DS_rotated}
\end{table*}

For subtasks A \& B, we experiment with FlanT5 zero-shot prompting, utilizing the Hugging Face Transformers\footnote{\url{huggingface.co/docs/transformers/}} library, specifically the T5ForConditionalGeneration class and T5Tokenizer. Training is conducted on an NVIDIA A100 GPU with 80GB memory over 24 hours. The prompting sample for subtask A is shown in Figure \ref{fig:prompt1}. In subtask B, we maintain consistency in prompting by keeping the question the same as labeling the human-generated text as "1", while prompting the machine-generated texts from various Language Model Models (LLMs) as categories "2" through "6."

We also finetune a t5-small model over 2 epochs, setting the learning rate to 0.001 and the batch size to 4. We employ a full finetuning (FFT) approach without the utilization of any quantization method like LoRa \cite{hu2021lora} or QLoRA \cite{dettmers2023qlora}. Due to the adoption of an FFT approach and the sheer size of the dataset, we do not experiment with a wide set of hyper-parameters. We empirically choose a few combinations and report the best results.

\section{Results}

Subtask A and B are evaluated based on Accuracy, as specified by \cite{semeval2024task8}, while Subtask C employs Mean Absolute Error (MAE) as the evaluation metric \footnote{\url{https://github.com/mbzuai-nlp/SemEval2024-task8}}.

 In the monolingual track of Subtask A, ELECTRA demonstrates superior accuracy (0.78) compared to RoBERTa (0.70) and DistilBERT (0.69) during the development phase. Consequently, the weighted ensemble of these three models achieves an accuracy of 0.79 in our development submission, surpassing the baseline RoBERTa model. Upon publishing test labels, a comparison with the test label results reveals accuracies detailed in Table \ref{tab:trackA_DS_rotated}, with the ensemble model achieving an accuracy of 0.74, while the baseline accuracy increases to 0.88, differing by 0.14 compared to the development phase. In the multilingual track, XLM-R outperforms LASER and mBERT with an accuracy of 0.61. Ensembling these models achieves accuracies of 0.63 in the development phase and 0.60 in the test phase, whereas the baseline accuracies are 0.72 and 0.81, respectively. Both zero-shot prompting and fine-tuning FlanT5 demonstrate less than satisfactory performance, yielding accuracies of 0.53 and 0.43 in the monolingual and multilingual tracks, respectively.

\begin{table}[!ht]
\centering
\begin{tabular}{lcc}
\hline
\textbf{Model} & \textbf{Dev} & \textbf{Test} \\
\hline
\bf Baseline (RoBERTa) & 0.75 & 0.75\\
\hline
FLAN-T5 Prompting & 0.54 & 0.48\\
FLAN-T5 Fine-tuning & 0.57 & 0.54\\
\hline
 RoBERTa   & 0.72 & 0.56\\
 ELECTRA  & 0.73 & 0.59\\
 DeBERTa   & 0.77 & 0.64\\
\hline
\bf Ensemble (Wt. accuracy) & 0.79 & 0.65\\
\hline
\end{tabular}
\caption{Accuracy of Multi-Way Machine-Generated Text Classification (Subtask B)}
\label{tab:B DS}
\end{table}

Within subtask B, DeBERTa outperforms RoBERTa and ELECTRA, achieving superior performance with an accuracy of 0.77. Ensembling these models yields accuracies of 0.79 and 0.65 in both the development and test phases, whereas baseline RoBERTa gives 0.75 in both phases. Similar to subtask A, fine-tuning and prompting FLAN T5 exhibit suboptimal results in both phases shown in Table \ref{tab:B DS}.

\begin{table}[!ht]
\centering
\begin{tabular}{lcc}
\hline
\textbf{Model} & \textbf{Dev} & \textbf{Test} \\
\hline
\bf Baseline (Longformer) & $\simeq$ 3.53 & 21.54\\
\hline
 TF-IDF + LR  & 44.15 & 71.23\\
 PPMI + LR & 41.93 & 68.41\\
RoBERTa + LR & 37.52 & 65.82\\
\hline
TF-IDF + EN  & 38.36 &67.09\\
PPMI + EN   & 35.67 & 63.36\\
RoBERTa + EN & 33.28 & 62.34\\
\hline
\bf Wt. (dev. MAE) Ensemble & 31.71 & 60.78\\
\hline
\end{tabular}
\caption{Mean Absolute Error(MAE) value of Human-Machine Mixed Text Detection (Subtask C) (LR = Linear Regression, EN = ElasticNet)}
\label{tab:C DS}
\end{table}

In subtask C, various methods are considered, and it is found that RoBERTa with Elastic Net achieved the minimum Mean Absolute Error (33.28). Table \ref{tab:C DS} highlights that Elastic Net outperforms Linear Regression in terms of lower MAE during both the development and test phases. To enhance predictive performance, we employ a weighted ensemble of development phase MAE of six combinations, resulting in MAE values of 31.71 and 60.78 during the development and test phases, respectively. However, the baseline (longformer) model gives MAE of $3.53 \pm 0.21$ and 21.54.

\section{Error Analysis}

In the monolingual track of Subtask A, the final model demonstrates proficiency in accurately identifying machine-generated text. Nonetheless, there is a notable presence of false positives, indicating instances where the model incorrectly identifies human-written texts as machine-generated. Despite this, the model effectively detects machine-generated text without omission. Similarly, in the multilingual track of Subtask A, the ultimate model excels in accurately distinguishing machine-generated text. However, false positives are prevalent, indicating numerous cases where human-written texts are inaccurately classified as machine-generated. Additionally, the model encounters instances where it fails to predict machine-generated texts.

In Subtask B, the model excels in accurately predicting chatGPT-generated texts. However, its performance declines notably for davinci-generated text, often misclassifying it as chatGPT generated. Additionally, the model's accuracy is lower for Dolly-generated and human-written texts, indicating a discrepancy in handling machine-generated versus human-written content.

For subtask C, MAE is higher due to the presence of outliers because the dev MAE was significantly lower than the test MAE. To handle this issue, it is essential to address the preprocessing of data, handling outliers, selecting appropriate features, optimizing model complexity, improving data quality, and ensuring model stability through proper tuning and evaluation procedures. This can be the future scope of research in this specific domain.

For a clearer understanding, refer to the visual evaluations in Figure \ref{fig:st1}, \ref{fig:st2}, \ref{fig:st3} of Appendix.

\section{Conclusion}

In our investigation of SemEval-2024 Task 8, we applied a diverse set of methodologies, encompassing statistical machine learning techniques, transformer-based models, sentence transformers, and FLAN T5. Subtask A involved binary classification, where the monolingual track focused on cross-generator scenarios within the same domain, and the multilingual track addressed cross-domain scenarios within the same generators. Subtask B dealt with multi-label classification, requiring the discrimination of human-generated text from five distinct language models. Subtask C centered on Human-Machine Mixed Text Detection, employing TF-IDF, PPMI, and RoBERTa with Linear Regression and ElasticNet for prediction.
The outcomes of three subtasks highlighted the efficacy of ensemble methods, showcasing specific models excelling in each subtask. Additionally, we explored the applicability of zero-shot prompting and fine-tuning FLAN-T5 for Tracks A and B.

In summary, our approach harnessed a blend of transformer models, machine learning methodologies, and ensemble strategies to tackle the complexities presented by SemEval-2024 Task 8. The paper underscores the imperative need for robust detection methods to effectively navigate the growing prevalence of machine-generated content.

\section*{Limitations}
The task involved extensive datasets in each phase of all subtasks, leading to prolonged execution times and increased GPU usage. Additionally, the texts themselves were lengthy. Moreover, the prohibition of additional data augmentation added to the complexity of the task. The nuanced distinction between human-written and machine-generated text, which can sometimes be challenging for humans to discern, poses an even greater difficulty for models attempting to learn this differentiation. %Exploring the potential of leveraging other up-to-date LLMs may show better performance in addressing these challenges.

\section*{Acknowledgements}

We express our gratitude to the organizers for orchestrating this task and to the individuals who diligently annotated datasets across various languages. Your dedication has played a crucial role in the triumph of this undertaking. %The meticulously designed task underscores the organizers' dedication to advancing research, and we commend the collaborative endeavors that have enhanced the diversity and comprehensiveness of the datasets, ensuring a substantial and positive influence.%
\bibliography{custom}

\appendix

\section{Appendix}

\begin{figure*}
  \centering
  \includegraphics[width=0.9\textwidth]{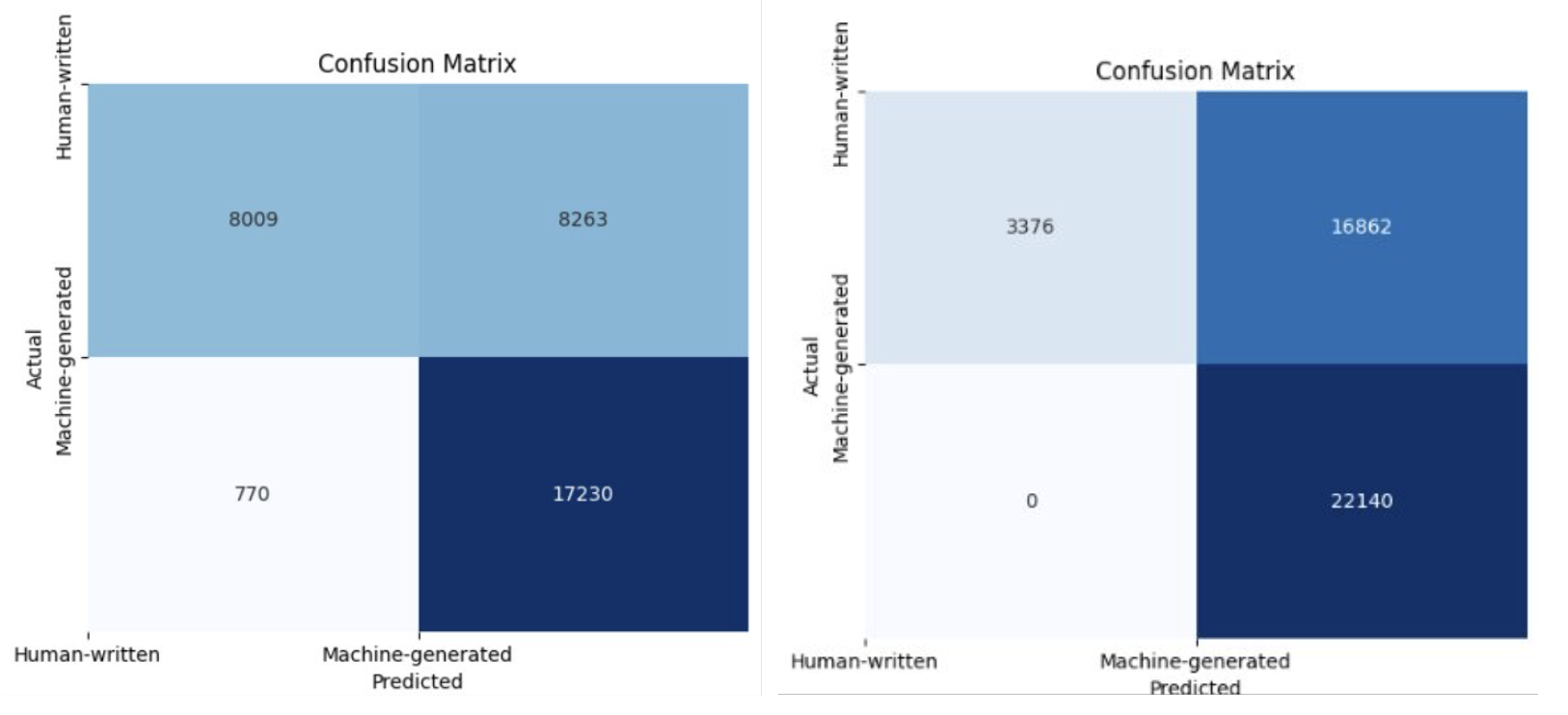}
  \caption{Confusion Matrix (Binary Human-Written vs. Machine-Generated Text Classification : Monolingual (Left), Multilingual (Right))}
  \label{fig:st1}
\end{figure*}

\begin{figure*}
  \centering
  \includegraphics[width=0.5\textwidth]{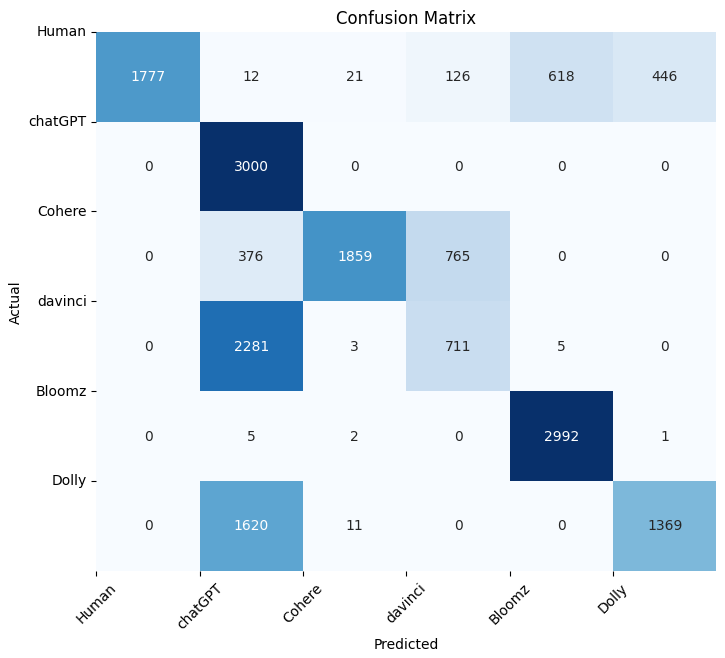}
  \caption{Confusion Matrix (Multi-Way Machine-Generated Text Classification)}
  \label{fig:st2}
\end{figure*}

\begin{figure*}
  \centering
  \includegraphics[width=0.6\textwidth]{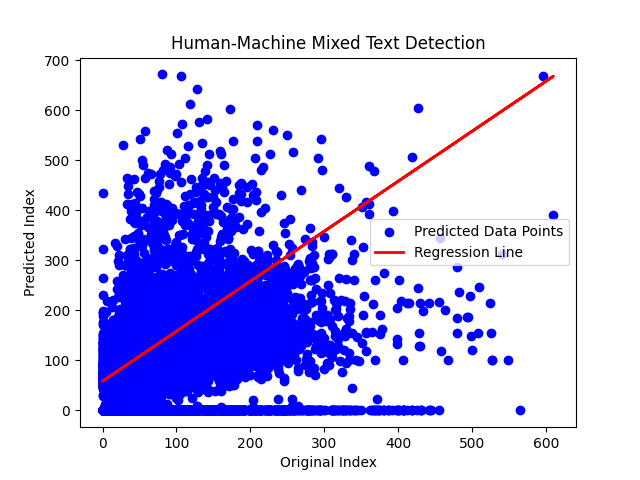}
  \caption{Regression (Human-Machine Mixed Text Detection)}
  \label{fig:st3}
\end{figure*}

\end{document}